\documentclass[a4paper,twoside]{article}

\usepackage{epsfig}
\usepackage{subcaption}
\usepackage{calc}
\usepackage{amssymb}
\usepackage{amstext}
\usepackage{amsmath}
\usepackage{amsthm}
\usepackage{multicol}
\usepackage{pslatex}
\usepackage{apalike}
\usepackage{dsfont}
\usepackage[bottom]{footmisc}
\usepackage{SCITEPRESS}     

\begin{document}

\title{Moving Other Way: Exploring Word Mover Distance Extensions\thanks{This work is an output of a research project implemented as part of the Basic Research Program at the National Research University Higher School of Economics (HSE University).}}

\author{\authorname{Ilya S. Smirnov\sup{1}, Ivan P. Yamshchikov\sup{1}\orcidAuthor{0000-0003-3784-0671}}
\affiliation{\sup{1}LEYA Lab, Yandex, Higher School of Economics}
}


\abstract{The word mover's distance (WMD) is a popular semantic similarity metric for two texts. This position paper studies several possible extensions of WMD. We experiment with the frequency of words in the corpus as a weighting factor and the geometry of the word vector space. We validate possible extensions of WMD on six document classification datasets. Some proposed extensions show better results in terms of the k-nearest neighbor classification error than WMD.}

\onecolumn \maketitle \normalsize \setcounter{footnote}{0} \vfill

\section{\uppercase{Introduction}}
\label{sec:introduction}

Semantic similarity metrics are essential for several Natural Language Processing (NLP) tasks. When working with paraphrasing, style transfer, topic modeling, and other NLP tasks, one usually has to estimate how close the meanings of two texts are to each other. The most straightforward approaches to measure the distance between documents rely on some scoring procedure for the overlapping bag of words (BOW) and term frequency-inverse document frequency (TF-IDF). However, such methods do not incorporate any semantic information about the words that comprise the text. Hence, these methods will evaluate documents with different but semantically similar words as entirely different texts. There are plenty of other metrics of semantic similarity: chrF \cite{popovic2015chrf} - character n-gram score that measures the number of n-grams that coincide both input and output; BLEU \cite{papineni2002bleu} developed for automatic evaluation of machine translation, or the BERT-score proposed in \cite{zhang2019bertscore} for estimation of text generation. In \cite{yamshchikov2020style} authors show that many current metrics of semantic similarity have significant flaws and do not entirely reflect human evaluations. At the same time, these evaluations themselves are pretty noisy and heavily depend on the crowd-sourcing procedure from \cite{solomon2021rethinking}. 

One of the most successful metrics in this regard is Word Mover's Distance (WMD) \cite{kusner2015word}. WMD represents text documents as a weighted point cloud of embedded words. It specifies the geometry of words in space using pretrained word embeddings and determines the distances between documents as the optimal transport distance between them. Many NLP tasks, such as document classification, topic modeling, or text style transfer, use WMD as a metric to automatically evaluate semantic similarity since it is reliable and easy to implement. It is also relatively cheap computationally and has an intuitive interpretation. For these reasons, this paper experiments with possible extensions of WMD. In this position paper, we discuss possible ways to improve WMD without losing its interpretability and without making the calculation too resource-intensive. 

We perform a series of experiments calculating WMD for different pretrained embeddings in vector spaces with different geometry. The list includes Hyperbolic space \cite{dhingra2018embedding} and tangent space of the probability simplex represented with Poincare embeddings \cite{nickel2017poincare,tifrea2018poincare} and Alpha embeddings \cite{volpi2021natural}. We suggest several word features that might affect WMD performance. We also discuss which directions seem to be the most promising for further metric improvements.

\section{\uppercase{Word Mover's Distance}}

Word Mover's Distance is fundamentally based on the Kantorovich problem between discrete measures, which is one of the fundamental problems of optimal transport (OT). Formally speaking, there should be several inputs to calculate the metric:
\begin{itemize}
    \item two discrete measures or probability distributions $\alpha$, $\beta$:
  
    \[\alpha = \sum_{i=1}^{n} \textbf{a}_i \delta_{x_i}, \beta = \sum_{i=1}^{m} \textbf{b}_i \delta_{y_i}\]
    
    where $x_1, \cdots, x_n \in R^d \; , \; y_1, \cdots , y_m \in R^d$, $\delta_x$ is the Dirac at position $x$, \textbf{a} and \textbf{b} are Dirac's weights.
    
    One also has a contract that $\sum_{i=1}^{n} \textbf{a}_i = \sum_{i=1}^{m} \textbf{b}_i$, and she tries to move the first measure to the second one so that they are equal.
    
    \item The transportation cost matrix $\textbf{C}$:
    
    \[\textbf{C}_{ij} = c(x_i, y_j)\]
    
    where $c(x_i, y_j): \mathds{R}^d \times \mathds{R}^d \to \mathds{R}$ is a distance or transportation cost between $x_i$ and $y_j$ 
\end{itemize}

Earth Mover's Distance or solution of the Kantorovich problem between $\alpha$ and $\beta$ is then defined through the following optimization problem:

\[EMD(\alpha, \beta, \textbf{C}) = \min\limits_{P \in \mathds{R^{n x m}}} \sum\limits_{i, j} \textbf{C}_{ij} \textbf{P}_{ij}\]

\[s.t. \;\;\;\;\; \textbf{P}_{ij} \geq 0, \; \textbf{P} \mathds{1} = \textbf{a}, \; \textbf{P}^T \mathds{1} = \textbf{b}\]

$\textbf{P}_{ij}$ intuitively represents the "amount" of word $i$ transported to word $j$.

Vanilla WMD is the cost of transporting a set of word vectors from the first document, represented as a bag of words, into a set of word vectors from the second document in a Euclidean space. So it is just an EMD but with some conditions:

\begin{itemize}
    \item probability distribution in terms of a document:

    \[\textbf{a} = \sum\limits_{i=1}^{n} a_i \delta_{w_i}\]
    \[s.t. \;\;\;\;\; \sum\limits_{i=1}^{n} a_i = 1\]

    where $w_1, \cdots, w_n$ is the set of words in a document, $a_i$  stands for the number of times the word $w_i$ appeared in the document divided by a total number of words in a document.
    
    \item $\textbf{C}(w_i, w_j) = \lVert w_i - w_j \rVert^2$
\end{itemize}

Now that we have described WMD in detail let us discuss several essential results that emerged since the original paper \cite{kusner2015word}, where it was introduced for the first time.

\section{\uppercase{Related Work}}

Word2Vec \cite{mikolov2013distributed} and GloVe \cite{pennington2014glove} are the most famous semantic embeddings of words based on their context and frequency of co-occurrence in text corpora. They leverage the so-called distributional hypothesis \cite{harris1954distributional}, which states that similar words tend to appear in similar contexts. Word2Vec and Glove vectors are shown to effectively capture semantic similarity at the word level. Word Mover's Distance takes this underlying word geometry into account but also utilizes the ideas of optimal transport and thus inherits specific theoretical properties from OT. It is continuously used and optimized for various tasks.

\cite{huang2016supervised} propose an efficient technique to learn a supervised WMD via leveraging semantic differences between individual words discovered during supervised training. \cite{yokoi2020word} demonstrate in their paper that Euclidean distance is not appropriate as a distance metric between word embeddings and use cosine similarity instead. They also weight documents' BOW with L2 norms of word embeddings. \cite{wang2020robust} replace assumption that documents' BOWs have the same measure to solve Kantorovich problem of optimal transport with the usage of Wasserstein-Fisher-Rao distance between documents based on unbalanced optimal transport principles. The work of \cite{sato2021re} is especially significant for further discussion. The authors re-evaluate the performances of WMD and the classical baselines and find that once the data gets L1 or L2 normalization, the performance of other classical semantic similarity measures becomes comparable with WMD. The authors also show that WMD performs better with TF-IDF regularization. In high-dimensional spaces, WMD behaves similarly to BOW, while in low-dimensional spaces, it seems to be influenced by the dimensionality curse.

\cite{zhao2019moverscore}, \cite{zhao2020limitations} and \cite{clark2019sentence} use WMD for different text generation applications, whereas \cite{lei2018discrete} use WMD to generate paraphrased texts. \cite{deudon2018learning} extends Word Mover’s Distance by representing text documents as normal distributions instead of bags of embedded words. \cite{chen2020improving} introduce fast-computed Tempered-WMD parameterized by temperature regularization inspired by Sinkhorn distance from \cite{cuturi2013sinkhorn}.
\cite{sun2018hierarchical} show that WMD performs quite well on a hierarchical multilevel structure.

\begin{table*}[h]
\vspace{-0.2cm}
\caption{Information about used datasets}\label{tab:example2} \centering
\begin{tabular}{|c|c|c|c|c|c|c|}
  \hline
   & twitter & imdb & amazon & classic & bbcsport & ohsumed \\
  \hline
  Total number of texts & 3115 & 1250 & 1500 & 2000 & 737 & 1250 \\
  Train number of texts & 2492 & 1000 & 1200 & 1600 & 589 & 1000 \\
  Test number of texts & 623 & 250 & 300 & 400 & 148 & 250 \\
  Average word's L2 norm & 2.60 & 2.76 & 2.77 & 2.86 & 2.78 & 3.03 \\
  Average text's length in words & 10.57 & 87.12 & 165.50 & 54.50 & 144.79 & 91.55 \\
  \hline
\end{tabular}
\end{table*}

\begin{table*}[h]
\vspace{-0.2cm}
\caption{kNN classification errors on all datasets for some variations of WMD. The results are reported on the best number of neighbors selected from $[1; 20]$}\label{tab:example2} \centering
\begin{tabular}{|c|c|c|c|c|c|c|}
  \hline
   & twitter & imdb & amazon & classic & bbcsport & ohsumed \\
  \hline
  WMD & 29.4 $\pm$ 1.7 & 24.8 & 9.7 $\pm$ 2.3 & 5.7 $\pm$ 1.9 & 3.4 $\pm$ 1.7 & 92.4 \\
  WMD-TF-IDF & 29.2 $\pm$ 1.0 & \bf 21.2 & 9.0 $\pm$ 1.7 & 4.9 $\pm$ 1.5 & 2.7 $\pm$ 1.0 & 92.0 \\
  WRD & \bf 28.7 $\pm$ 1.3 & 23.2 & \bf 7.2 $\pm$ 2.1 & \bf 4.1 $\pm$ 1.3 & 3.6 $\pm$ 1.1 & 90 \\
  $OPT_1$ & 29.6 $\pm$ 1.2 & 27.2 & 20.7 $\pm$ 4.0 & 15.2 $\pm$ 12.9 & 4.1 $\pm$ 1.5 & \bf 88.5 \\
  $OPT_2$ & 29.6 $\pm$ 1.5 & 27.2 & 10.1 $\pm$ 1.5 & 5.8 $\pm$ 1.8 & \bf 2.6 $\pm$ 1.1 & 92.0 \\
  \hline
\end{tabular}
\end{table*}

\section{\uppercase{Experimental settings}}

This section describes the experiments that we carry out in detail.

\subsection{Datasets}

To assure better reproducibility, we work with the datasets presented in \cite{kusner2015word} and \cite{sato2021re}\footnote{The data is available online https://github.com/mkusner/wmd}. For the evaluation, we use six datasets that we believe to be diverse and illustrative enough for the aims of this discussion. The datasets are TWITTER, IMDB, AMAZON, CLASSIC, BBC SPORT, and OHSUMED.

We remove stop words from all datasets, except TWITTER, as in the original paper \cite{kusner2015word}. IMDB and OHSUMED datasets have predefined train/test splits. On four other datasets, we use 5-fold cross-validation. Due to time constraints to speed up the computations, we take subsamples from more extensive datasets. Table 1 shows the parameters of all the resulting datasets we use for the evaluation and experiments.

Similar to \cite{sato2021re}, we split the training set into an 80/20 train/validation set and select the neighborhood size from [1, 20] using the validation dataset. 

\subsection{Embeddings}

Since WMD relies on some form of word embeddings, we experiment with several pre-trained models and train several others ourselves. First, we use original 300-dimensional Word2Vec embeddings trained on the Google News corpus that contains about 100 billion  words\footnote{https://code.google.com/archive/p/word2vec/}. A series of works hint that original Euclidian geometry might be suboptimal for the space of word embeddings.

\cite{nickel2017poincare,tifrea2018poincare} suggest the Poincare embeddings that map words in a hyperbolic rather than a Euclidian space. Hyperbolic space is a non-Euclidean geometric space with an alternative axiom instead of Euclid's parallel postulate. In hyperbolic space, circle circumference and disc area grow exponentially with radius, but in Euclidean space, they grow linearly and quadratically, respectively. This property makes hyperbolic spaces particularly efficient to embed hierarchical structures like trees, where the number of nodes grows exponentially with depth. The preferable way to model Hyperbolic space is the Poincare unit ball, so all embeddings $v$ will have $\lVert v \rVert \leq 1$. Poincare embeddings are learned using a loss function that minimizes the hyperbolic distance between embeddings of similar words and maximizes the hyperbolic distance between embeddings of different words. 

\cite{volpi2021natural} proposes alpha embeddings as a generalization to the Riemannian case where the computation of the cosine product between two tangent vectors is used to estimate semantic similarity. According to Information Geometry, a statistical model can be modeled as a Riemannian manifold with the Fisher information matrix and a family of $\alpha$ connections. Authors propose a conditional SkipGram model that represents an exponential
family in the simplex, parameterized by two matrices $U$ and $V$ of size $n \times d$, where $n$ is the cardinality of the dictionary, and $d$ is the size of the embeddings. Columns of V determine the sufficient statistics of the model, while each row $u_w$ of $U$ identifies a
probability distribution. The alpha embeddings are defined up to the choice of a reference distribution $p_0$.The natural alpha embedding of a given word $w$ is defined as the projection of the logarithmic map onto the tangent space of some submodel. We will not dive into more detail since this is beyond the scope of our work but address the reader to the original paper \cite{volpi2021natural} .

We train Poincare embeddings for Word2Vec and alpha embeddings over GloVe in 8 different dimensionalities on text8 corpus\footnote{https://deepai.org/dataset/text8} that contains around 17 million words. The SkipGram models for both Word2Vec and Poincare embeddings are trained with similar parameters: the minimum number of occurrences of a word in the corpus is 50, the size of the context window is 8, negative sampling with 20 samples. The number of epochs is five, and the learning rate decreases from 0.025 to 0. 

Similar to \cite{volpi2021natural}  we use GloVe embeddings as a base for alpha embeddings and train it for fifteen epochs. The word2vec Skip-Gram with negative sampling is equivalent to a matrix factorization with GloVe so it is easy to reproduce using the original framework\footnote{https://github.com/rist-ro/argo}. The co-occurrence matrix is built with the minimum number of occurrences of a word in the corpus being 50 and the window size equal to 8.

\subsection{Models}

We use only vanilla WMD to compare embedding in different spaces, but the distance between word embeddings is measured differently depending on the geometry of the underlying space. We set hyperparameter $\alpha$ in tangent space of the probability simplex equal to 1 to ease distance computation.

\begin{itemize}
    \item Euclidean space
    \[c(w_i, w_j) = \lVert w_i - w_j \rVert^2 \]
    
\begin{table*}[h]
\vspace{-0.2cm}
\caption{kNN classification errors on TWITTER dataset for all embeddings' types and different embeddings' dimensions. The best number of neighbors was selected from the segment $[1; 20]$}\label{tab:example2} \centering
\begin{tabular}{|c|c|c|c|}
  \hline
   & \multicolumn{1}{|p{2cm}|}{\centering Word2Vec \\ embeddings} & \multicolumn{1}{|p{2cm}|}{\centering Poincare \\ embeddings} & \multicolumn{1}{|p{2cm}|}{\centering Alpha \\ embeddings}\\
  \hline
  5 & \bf 33.1 $\pm$ 2.7  & 35.0 $\pm$ 3.7 & 36.4 $\pm$ 4.3 \\
  10 & \bf 33.2 $\pm$ 3.3 & 36.7 $\pm$ 5.3 & 36.5 $\pm$ 4.7 \\
  25 & \bf 33.8 $\pm$ 2.8 & 34.7 $\pm$ 3.5 & 35.4 $\pm$ 4.5 \\
  50 & \bf 33.5 $\pm$ 2.5 & 37.6 $\pm$ 6.4 & 37.1 $\pm$ 6.5 \\
  100 & 34.4 $\pm$ 3.2 & \bf 33.2 $\pm$ 2.9 & 38.2 $\pm$ 6.7 \\
  200 & \bf 34.4 $\pm$ 3.2 & 34.7 $\pm$ 3.4 & 35.5 $\pm$ 4.0 \\
  300 & 34.5 $\pm$ 3.4 & 36.6 $\pm$ 4.2 & \bf 33.1 $\pm$ 2.4 \\
  400 & 34.9 $\pm$ 3.8 & 34.4 $\pm$ 3.6 & \bf 34.4 $\pm$ 1.9 \\
  \hline
\end{tabular}
\end{table*}

\begin{table*}[h]
\vspace{-0.2cm}
\caption{kNN classification errors on IMDB dataset for all embeddings' types and different embeddings' dimensions. The best number
of neighbors was selected from the segment $[1; 20]$}\label{tab:example2} \centering
\begin{tabular}{|c|c|c|c|c|c|c|c|c|}
  \hline
   & 5 & 10 & 25 & 50 & 100 & 200 & 300 & 400 \\
  \hline
  Word2Vec embeddings & 43 & \bf 35 & \bf 29 & \bf 30 & \bf 24 & \bf 30 & \bf 28 & \bf 28\\
  Poincare embeddings & \bf 39 & 43 & 42 & 50 & 45 & 45 & 41 & 41\\
  Alpha embeddings & 52 & 48 & 63 & 49 & 52 & 48 & 57 & 55 \\
  \hline
\end{tabular}
\end{table*}

    \item Hyperbolic space (Unit Poincare ball)
    \[c(w_i, w_j) = \cosh^{-1} \left(1 + 2 \frac{\lVert w_i - w_j \rVert^2}{(1 - \lVert w_i \rVert^2)(1 - \lVert w_j \rVert^2)}\right)\]
    \item Tangent space of the probability simplex
    \[c(w_i, w_j) = \frac{w_i^T I(p_u) w_j}{\lVert w_i \rVert_{I(p_u)} \lVert w_j \rVert_{I(p_u)}}\]
    where $I(p_u)$ is the Fisher information matrix which could be computed during training alpha embeddings
\end{itemize}

To compare WMD variations on pretrained word embeddings, we also compare five variations of WMD. The main idea of the WMD variants that we experiment with is that one wants to prioritize the transportation of rare words. Naturally, the semantics of a rare word might carry far more meaning than the several frequently used ones. According to \cite{arefyev2018much} the embedding norm of a word positively correlates with its frequency in the training corpus. We use this idea and propose the following WMD variations for comparison:

\begin{itemize}
    \item vanilla WMD
    \item WMD with TF-IDF regularization applied to bags of words for both documents \cite{sato2021re}
    \item WRD - Word Rotator's Distance \cite{yokoi2020word}
    
    to compute it authors use \[1 - cos(w_i, w_j)\] as a distance or transportation cost between words $w_i$ and $w_j$.   They multiply the document's BOW by words norm. More precisely, let a document have $N$ unique words and $A = a_1, \cdots, a_N$, where $a_i$ is a number of times a word $w_i$ occurs in the document. New BOW is calculated like this: 
    \[A' = a_1 \cdot \lVert w_1 \rVert, \cdots, a_n \cdot \lVert w_n \rVert\]
    \item $OPT_1$: after calculating vanilla WMD between two documents, which have BOWs named as $A$ and $B$, we normalize the WMD score by the following coefficient:
    \[coeff = 1 + \sum\limits_{w_a = w_b} \frac{min(a, b)}{\lVert w_a \rVert^2}\]
    where $w_a$ and $a$ are a word and its frequency in the first document respectively, while $w_b$ and $b$ stand for the word and its frequency in the second one . 
    
    This coefficient makes WMD lower if there are rare matching words in both documents.
    
    \item $OPT_2$: we want to increase the measure of rare words relative to the rest of the words. So let's use rebalanced BOW with the formula inspired by TF-IDF: 
    
    \[A' = a_1 \cdot \log\left(\frac{d}{\lVert w_1 \rVert}\right), \cdots, a_n \cdot \log\left(\frac{d}{\lVert w_n \rVert}\right)\]
    
    where $d$ is the dimensionality of word embeddings.
    
    There is a simple idea behind this: the norm of rare words is less than that of the frequently used ones, $\log\left(\frac{d}{\lVert w \rVert}\right)$. Thus we increase the impact of rare words more while decreasing the effects of the frequent ones less.
\end{itemize}

\subsection{Evaluation and results}

Table 2 shows that overall our variations of WMD could behave quite badly. $OPT_1$ with a simple division of the final metric by a coefficient is especially crude when comparing it on all datasets. However, on some of the tasks, the proposed measures are either the best or close to the best result. 

So $OPT_1$ shows the best performance on the OHSUMED dataset, which contains medical abstracts categorized by different disease groups. This dataset has an abundance of rare words, thus it seems that the proposed normalization was useful because of this property of the data. Its bad performance on other datasets could be due to an excessive amount of frequent word matches in those documents.  

Looking at Tables 3 and 4, one can notice that on both datasets WMD with Word2Vec embeddings performs well and beats WMD with other embeddings. However, there are some outliers. On the TWITTER dataset, alpha embeddings perform best for the standard standard dimension of 300, which may signal the possible benefits of further studying them and learning or iterating over the hyperparameter $\alpha$. 

For embeddings of dimension 5 on the IMDB dataset, Poincare embeddings perform the best. Thus one could suggest that they capture semantics in low-dimensional spaces better than other embeddings' types.

We can also notice that on TWITTER the classification error is almost the same, whereas on IMDB the differences are noticeable. It seems that Poincare and Alpha embeddings better reflect the semantics of frequently used words.

\section{\uppercase{Discussion}}
\label{sec:discussion}
We want to point out some interesting moments for future research.

\textbf{Geometry of the underlying space.}\\
The Euclidean embedding space must be of large dimension. Other geometries show better results at lower dimensions. However, we are experimenting with small samples of two datasets. It would be interesting to check whether the superiority trend of Word2Vec embeddings in terms of WMD continues on embeddings of large dimensions for other datasets or larger subsamples of TWITTER and IMDB datasets.

\textbf{Normalization with word frequencies.}\\
The frequency of words in the training corpus affects the WMD score, but we make only several attempts to use it. This seems to be a promising direction for future research. Indeed, on the specialized datasets the variants that take word frequencies into account show good results.

\section{\uppercase{Conclusions}}
\label{sec:conclusion}
This position paper conducts a series of experiments to calculate Word Mover's Distance in different embedding spaces. 

It seems that taking into account the frequency of words and improving the mechanism of optimal transport in application to semantics could be promising directions for further research. However, additional work on this problem is necessary. 

Further, new embedding types have been found that behave well on specific dimensions, and further study of these embeddings can be meaningful within the framework of the semantic similarity problem. 
\bibliographystyle{apalike}
{\small
\bibliography{example}}

\end{document}